\documentclass{bmvc2k}

\title{Mitigating Hallucinations in Multimodal LLMs via Object-aware Preference Optimization}

\addauthor{Alberto Compagnoni}{alberto.compagnoni@unimore.it}{1}
\addauthor{Davide Caffagni}{davide.caffagni@unimore.it}{1}
\addauthor{Nicholas Moratelli}{nicholas.moratelli@unimore.it}{1}
\addauthor{Lorenzo Baraldi}{lorenzo.baraldi@unimore.it}{1}
\addauthor{Marcella Cornia}{marcella.cornia@unimore.it}{1}
\addauthor{Rita Cucchiara}{rita.cucchiara@unimore.it}{1}

\addinstitution{
 University of Modena and Reggio Emilia\\
 Modena, Italy
}

\runninghead{A. Compagnoni \etal}{Mitigating Hallucinations in Multimodal LLMs}

\usepackage{booktabs}       
\usepackage{amsfonts}       
\usepackage{nicefrac}       
\usepackage{microtype}      
\usepackage{xcolor}         
\usepackage{graphicx}
\usepackage{amsmath}
\usepackage{amssymb}
\usepackage{array}
\usepackage{multirow}
\usepackage{color}
\usepackage{colortbl}
\usepackage{framed}
\usepackage{bm}
\usepackage{xspace}
\usepackage{enumitem}
\usepackage{xparse}
\usepackage{algorithm}
\usepackage{listings}
\usepackage{url}
\usepackage{mathtools}
\usepackage{pifont}
\usepackage{mathtools}
\usepackage{lipsum}
\usepackage{adjustbox}
\usepackage{stfloats}
\usepackage{listings}

\newcommand{\cmark}{\ding{51}}%
\newcommand{\xmark}{\ding{55}}%

\definecolor{Gray}{gray}{0.2}
\definecolor{lightgray}{gray}{0.92}
\definecolor{OurColor}{rgb}{0.886, 0.941, 0.851}

\definecolor{OurRed}{HTML}{FD2121}
\definecolor{OurBlue}{HTML}{0070C0}

\definecolor{customgray}{gray}{0.35}

\newcommand{\tit}[1]{\smallbreak\noindent\textbf{#1.}}
\newcommand{\tinytit}[1]{\noindent\textbf{#1.}}

\newcommand{\ours}{CHAIR-DPO\xspace}

\newcommand{\pitheta}{\pi_{\theta}}
\newcommand{\piref}{\pi_\text{ref}}
\newcommand{\chair}{\text{CHAIR}}

\def\ie{\emph{i.e}\bmvaOneDot}
\def\eg{\emph{e.g}\bmvaOneDot}

\def\etal{\emph{et al}\bmvaOneDot}

\begin{document}

\maketitle

\begin{abstract}
Multimodal Large Language Models (MLLMs) emerge as a unified interface to address a multitude of tasks, ranging from NLP to computer vision. Despite showcasing state-of-the-art results in many benchmarks, a long-standing issue is the tendency of MLLMs to hallucinate, that is to generate answers to the user's query that are not reflected in the visual input. In this paper, we address the problem of hallucinations as an alignment problem, seeking to steer the MLLM so that it prefers generating content without hallucinations. In contrast to recent approaches that require complicated pipelines to build synthetic preference data for alignment training, often relying on proprietary models, we capitalize on the well-known CHAIR metric, originally proposed to gauge the degree of hallucinations in image captioning. Given a pair of generated answers, we leverage CHAIR to distinguish winner and loser options (\ie, non-hallucinated and hallucinated samples) and fine-tune off-the-shelf MLLMs via Direct Preference Optimization (DPO). The resulting method, which we refer to as \ours, effectively diminishes the amount of hallucinated answers on several hallucination benchmarks, demonstrating the effectiveness of fine-tuning the MLLM with a CHAIR-based reward. Source code and trained models are publicly available at \url{https://github.com/aimagelab/CHAIR-DPO}.
\end{abstract}


\section{Introduction}
\label{sec:intro}
Research interest in Multimodal Large Language Models (MLLMs) is raging. By capitalizing on massive self-supervised pre-training on text, images~\cite{liu2023visual,liu2023improved,bai2023qwenvl,ye2024mplug,laurenccon2024building}, and possibly other modalities~\cite{panagopoulou2023x,han2024onellm,sun2024generative}, they emerge as a unified interface the user can interact with to solve different problems~\cite{caffagni2024r}. Not only they are accessible to even non-expert users, as the interaction happens via natural language, but they deliver strong performance on several tasks, often challenging specialized models tailored to a unique task alone. The capabilities of an MLLM goes beyond the production of text in response to a user query, spanning from visual grounding and referring~\cite{peng2023kosmos,chen2023shikra,rasheed2024glamm}, up to the generation of images and videos~\cite{sun2023generative,sun2024generative,team2024chameleon}.

Despite their impressive capabilities, MLLMs still suffer from hallucinations -- generating content that is unsupported by the input. Hallucination is a long-standing problem widely studied in the natural language processing community~\cite{xu2024hallucination,huang2025survey}, but having access to complementary data modalities other than text stretches it out and opens new paths for the model to hallucinate~\cite{sahoo2024comprehensive}. For instance, visual hallucinations~\cite{rohrbach2018object,wang2023amber,huang2024opera,yue2024less,yin2024woodpecker} manifest whenever an MLLM mentions an object not depicted in the input image. 

\begin{figure*}[t]
\includegraphics[width=\linewidth]{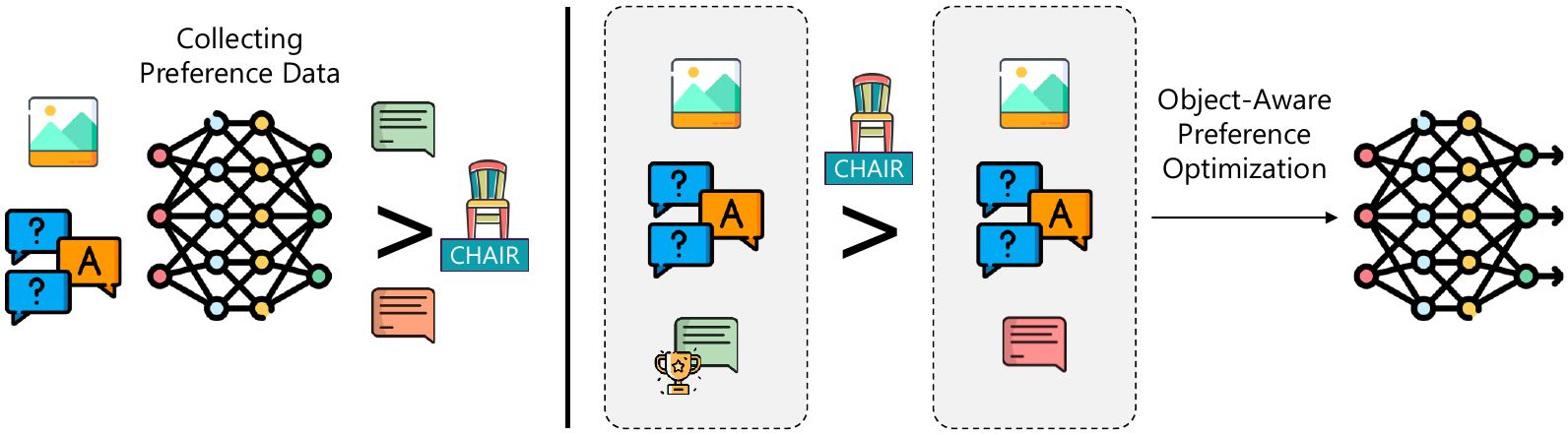}
\vspace{-0.4cm}
\caption{Overview of \ours. We begin by collecting preference data for optimization. Given two completions for a visual and textual prompt, we leverage $\chair_i$ to rank them: the preferred and dispreferred answers are chosen based on the lowest and highest number of hallucinated objects. Next, we use these preference pairs to fine-tune an MLLM with DPO, resulting in an object-aware model conscious of the objects truly depicted in the image.}
\vspace{-0.3cm}
\label{fig:method}
\end{figure*}

An intuitive framework for reducing hallucinations is to view them as a human alignment problem: as we humans reasonably prefer answers devoid of hallucinations, so should an MLLM properly aligned to human preference. Unfortunately, established techniques employed in developing MLLMs, such as visual instruction tuning~\cite{liu2023visual,liu2023improved}, Reinforcement Learning from Human Feedback (RLHF)~\cite{ouyang2022training}, or Direct Preference Optimization (DPO)~\cite{rafailov2023direct}, often prioritize that the generated answer effectively fulfills the user query, overlooking whether it contains hallucinations or not. Since they are implemented as the last training stage of MLLMs, they play the most important role in aligning the model to human preference.

Recent research~\cite{zhao2023beyond,wang2024mdpo,jiang2024modality,sarkar2025mitigating,wu2025generate} has introduced alignment methods to steer MLLMs toward preferring non-hallucinated outputs, with a particular focus on DPO. DPO is especially appealing as it replaces the complexity of reinforcement learning with a more tractable supervised learning approach. A key challenge prior to its application, however, is to collect preference data about hallucinations. In other words, how can we know which answer generated by an MLLM is hallucinated and which is not? Resorting to human annotators is costly and does not scale, so a compelling alternative becomes to query cutting-edge proprietary MLLMs such as GPT-4~\cite{zhao2023beyond,zhou2024aligning,wu2025generate} and Gemini~\cite{sarkar2025mitigating} to act as a judge. 

In this work, we introduce \ours, a new preference optimization method to tackle visual hallucinations in MLLMs. \ours builds on top of $\chair$~\cite{rohrbach2018object}, a well-known metric proposed to assess the degree of hallucination of image captioning models~\cite{sarto2025image,barraco2023little,petryk2024aloha}. The key idea is to leverage $\chair$ to quantitatively measure hallucinations in the responses generated by an MLLM, and thus select as preferred the response with the lower hallucination rate. After collecting enough preference pairs, we apply DPO to fine-tune an MLLM, enhancing its awareness concerning the presence or absence of objects in the input image, and drastically reducing visual hallucinations. An overview of the proposed approach is outlined in Figure~\ref{fig:method}.

We argue that $\chair$ enables more efficient preference data collection for hallucination mitigation compared to existing approaches. Experiments on multiple hallucination benchmarks, such as AMBER~\cite{wang2023amber}, CHAIR-MSCOCO~\cite{yue2024less}, and Object HalBench~\cite{rohrbach2018object, yu2024rlhf}, show that \ours achieves state-of-the-art performance, significantly reducing hallucination rates without degrading the original capabilities of the underlying MLLM.

\section{Related Work}
\label{sec:related}

\tinytit{Multimodal Large Language Models}
Building on the advancements of Large Language Models (LLMs), interest is surging to extend LLMs to the multimodal domain~\cite{liu2023improved,bai2023qwenvl,sun2023generative,chen2023shikra,ye2024mplug,laurenccon2024building,sun2024generative,rasheed2024glamm,team2024chameleon}, with most of the effort devoted to the integration of visual understanding. MLLMs perceive different modalities thanks to external encoders (\eg, CLIP visual encoder~\cite{radford2021learning} for images) that transform (unimodal) data into modality-specific embeddings that can be understood by the LLM upon the application of an adapter module. In this work, we utilize open-source MLLMs~\cite{liu2023improved,cocchi2025llavamore} built on the popular LLaVA~\cite{liu2023visual,liu2023improved} framework. These models feature a CLIP-ViT-L/14@336~\cite{radford2021learning} as the visual encoder and a lightweight feed-forward network as the vision-to-language adapter. LLaVA initially trains only the adapter module on an image captioning dataset. Next, it also unfreezes the pre-trained LLM and jointly optimizes it on a visual instruction tuning dataset comprising multi-turn visual dialogues. 
The parameters of the CLIP visual encoder are kept frozen and never updated.

\tit{Direct Preference Optimization (DPO)} DPO~\cite{rafailov2023direct} has been established as a compelling alternative to Reinforcement Learning from Human Feedback (RLHF)~\cite{ouyang2022training} to align fine-tuned LLMs with human preferences. DPO presents two main advantages against RLHF. First, it saves the effort of training a reward model to mimic human preferences, and second, it completely bypasses the reinforcement learning stage, which is non-trivial to implement in practice. The intuition behind DPO is to frame the constrained minimization problem of RLHF such that the objective does not rely on the reward gained by the model anymore, but instead, it depends upon an implicit reward expressed in terms of the optimizing and reference policies. Originally born for LLMs only, DPO and its variants~\cite{song2024preference, hong2024orpo, wu2024beta} are now successfully applied for aligning MLLMs as well~\cite{jia2024symdpo, zhang2024direct}, not to mention different generative domains spanning from diffusion models~\cite{wallace2024diffusion} to image captioning~\cite{moratelli2024revisiting}. 

\tit{Hallucination Mitigation} 
Despite the success of MLLMs, hallucinations remain an open challenge. Recent research~\cite{bai2024hallucination} identifies the root of hallucinations in noisy or corrupted training samples, eventually exacerbated by the addition of synthetic training data, as well as an imbalance in the frequency of certain entities appearing more often than others during optimization. This is further compounded by the overwhelming dominance of the LLM compared to the size of the visual encoder, which may lead to an over-reliance on linguistic priors rather than visual grounding. To mitigate hallucinations, prior work explores training-free and training-based approaches. Training-free methods~\cite{chuang2024dola,leng2024mitigating,huang2024opera} modify decoding strategies or apply post-hoc corrections~\cite{yin2024woodpecker} using external models. Training-based approaches fine-tune MLLMs via variants of preference alignment losses~\cite{zhao2023beyond,wang2024mdpo,jiang2024modality,zhou2024aligning,sarkar2025mitigating} or modified cross-entropy objectives~\cite{yue2024less,wu2025generate}. Notably, HA-DPO~\cite{zhao2023beyond} augments DPO with auxiliary language modeling loss component, while HALVA~\cite{sarkar2025mitigating} uses a fine-grained feedback to only penalize hallucinated tokens at phrase level. Lastly, both mDPO~\cite{wang2024mdpo} and MFPO~\cite{jiang2024modality} incorporate an image preference loss and anchor terms to preserve visual grounding and prevent reward degradation. Crucially, most of these approaches create or use datasets derived from complicated pipelines, often leveraging proprietary MLLMs. In contrast, our method relies solely on (i) an open-source model, the same one being fine-tuned, used exclusively for sampling completions, and (ii) an off-the-shelf object detector, drastically simplifying preference data collection.

\section{Proposed Method}
\label{sec:method}

\begin{figure*}[t]
\includegraphics[width=\linewidth]{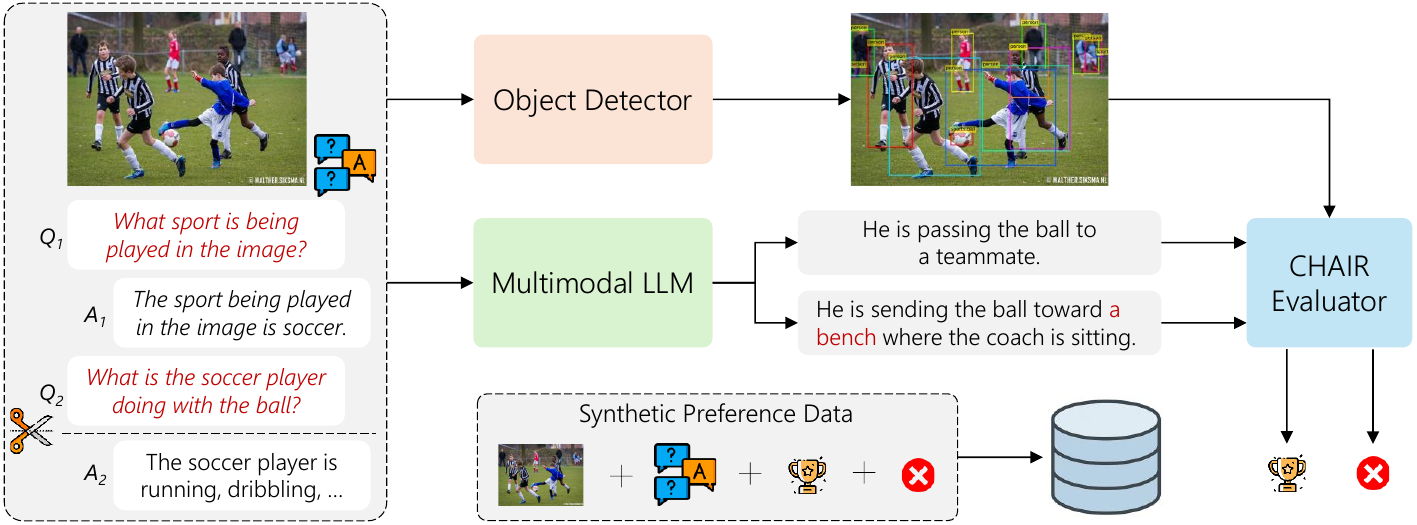}
\vspace{-0.4cm}
\caption{Details of the preference data collection pipeline. First, we ask an MLLM to generate two completions conditioned on a textual prompt $x_T$ and an input image $x_I$. Next, we use a pre-trained object detector to obtain the set of class names of the objects appearing in $x_I$. We use this set to compute $\chair_i$ to evaluate the completions: lower $\chair_i$ score identifies the preferred one, as it contains fewer hallucinations.}
\label{fig:data_collection}
\vspace{-0.3cm}
\end{figure*}

\subsection{Problem Formulation}
In our settings, an MLLM conditioned on textual prompt $x_T$ and image $x_I$ outputs the continuation text $y$ with probability $P(y \mid x_T, x_I)$. We call $y$ a hallucinated answer and denote it as $y_l$, whenever $y$ mentions any object not present in $x_I$. Building on the intuition that humans should prefer a non-hallucinated answer $y_w$ more than a hallucinated one, we seek to align the model with this preference, so that the probability of generating $y_w$ gets higher than $y_l$.

\subsection{Collecting Preference Data for Object Hallucinations} \label{ssec:data_collection}
To align our model with human preferences regarding object hallucinations, we need a supervised signal that can distinguish between non-hallucinated and hallucinated responses. In this work, we create this signal with CHAIR~\cite{rohrbach2018object}, a metric designed to assess the hallucination extent of image captioning models. We summarize in Figure~\ref{fig:data_collection} our data collection process. Given an answer $y$ generated by an MLLM, we define its $\chair_i$ score as the fraction of hallucinated object instances mentioned in $y$:
\begin{equation}
    \chair_i(y) = \frac{\left | \text{[hallucinated objects]}_y \right |}{\left | \text{[all mentioned objects]}_y \right |}.
    \label{eq:chair_i}
\end{equation}
Note that the hallucinated and mentioned objects in Eq.~\ref{eq:chair_i} are not treated as a set, but rather as a \textit{list}. It follows that $\chair_i$ can penalize a sentence for hallucinating the same objects multiple times. Normally, $\chair_i$ is applied on a supervised dataset where the objects mentioned in a ground-truth caption are known in advance. As we do not have access to similar annotations concerning hallucination, we propose to adapt publicly available datasets commonly used for visual instruction tuning of MLLMs~\cite{liu2023improved}. Such a dataset can be seen as a collection of triplets $\{x_T, x_I, y\}$, where $y$ is the ground-truth answer conditioned on the prompt $x_T$ and image $x_I$. In case of multi-turn dialogs, we randomly truncate the conversation, and provide the model with prior turns along with the current question as textual context.

First of all, for each triplet in the dataset, we discard $y$, and sample a new pair of possible answers $\{y_1, y_2\}$ from a reference instruction-tuned MLLM $\piref$. We then apply an off-the-shelf detector trained to recognize a predefined set of objects to the image $x_I$, treating the class name of the detected items as the ground-truth set of objects present in $x_I$. At this point, it is possible to compute the $\chair_i$ score of the generated answers, considering as hallucinated any mentioned object that is not in the ground-truth set. To make the evaluation robust, we leverage a predefined set of synonyms to match the words in the generated answers with the class names of the ground-truth objects.
We can now designate the non-hallucinated (\ie, winner) and hallucinated (\ie, loser) answers as:
\begin{equation}\label{eq:chair_pref}
y_w = \min_{y \in \{y_1, y_2\}} \text{CHAIR}_i(y), \quad y_l = \max_{y \in \{y_1, y_2\}} \text{CHAIR}_i(y),
\quad \text{with}~y_1, y_2 \sim \piref(y \mid x_T, x_I).
\end{equation}

\subsection{Object-Aware Preference Optimization}
With our preference data established, we apply Direct Preference Optimization (DPO)~\cite{rafailov2023direct}, an effective training approach for the human alignment of MLLMs. DPO bypasses the need for reinforcement learning to explicitly maximize a reward signal that scores the completions generated by the model. Instead, DPO simultaneously trains the policy model (\ie, the chosen MLLM) along with an implicit reward model that assigns a higher score to a winner answer $y_w$ compared to the loser answer $y_l$:
\begin{equation}
\label{eq:dpo}
\resizebox{0.95\textwidth}{!}{$
    \mathcal{L}_\text{DPO}(\pitheta; \pi_\text{ref}) = -\mathbb{E}{(x_T, x_I, y_w, y_l)\sim \mathcal{D}}\left[\log \sigma \left(\beta \log \frac{\pi_{\theta}(y_w\mid x_T, x_I)}{\pi_\text{ref}(y_w\mid x_T, x_I)} -\beta \log \frac{\pi_{\theta}(y_l\mid x_T, x_I)}{\pi_\text{ref}(y_l\mid x_T, x_I)}\right)\right],
    $}
\end{equation}
where $\beta$ is a regularizing hyperparameter controlling the strength of the Kullback-Leibler divergence, expressing the degree to which the policy model $\pitheta$ must be tied to a frozen reference model $\piref$. In practice, $\pitheta$ is initialized with the same weights as $\piref$. Note that our method may be applied iteratively: after exhausting the dataset once, fresh preference data can be built following Sec.~\ref{ssec:data_collection}, upon updating $\piref$ with $\pitheta$.

Minimizing Eq.~\ref{eq:dpo} pulls up the probability of generating the non-hallucinated completion $\pitheta(y_w \mid x_T, x_I)$ to the detriment of the hallucinated one $\pitheta(y_l \mid x_T, x_I)$. Succeeding in that has the immediate consequence of making $\pitheta$ \textit{aware} of what objects really appear in the presented image, and what are missing instead. We refer to the proposed method as \ours. 

\tit{Data Filtering for Reliable Preference Supervision}
A critical challenge we observe during training stems from the presence of several completion pairs with indistinguishable $\chair_i$ scores, making it impossible to assign winner and loser labels. Incorporating such pairs into the optimization process introduces noisy supervision, as the model is forced to learn from randomly chosen preference labels. To address this, we introduce a simple yet effective filtering strategy: we discard all training instances where the $\chair_i$ score difference between completions is zero. This ensures that the retained supervision pairs reflect a meaningful distinction in object hallucination severity, leading to a more reliable optimization signal.

\section{Experiments}
\label{sec:experiments}

\subsection{Experimental Setting}
\tinytit{Implementation and Training Details}
We build hallucination preference data starting from LLaVA-Instruct-665k~\cite{liu2023improved}, an open-source and widely employed dataset for visual instruction tuning, comprising multi-turn dialogues. We ask our chosen MLLMs, namely LLaVA-1.5-7B~\cite{liu2023improved} and LLaVA-MORE-8B~\cite{cocchi2025llavamore}\footnote{Specifically, LLaVA-1.5-7B is based on Vicuna-7B~\cite{vicuna2023}, while LLaVA-MORE-8B is based on LLaMA-3.1-8B~\cite{dubey2024llama}. Both are trained following the two-stage training pipeline introduced in~\cite{liu2023improved}.}, to generate the candidate completions $\{y_1, y_2\}$ conditioned on the image and the first $k$ human-assistant turns, where $k$ is randomly chosen. Next, we identify winner and loser candidates by computing the $\chair_i$ score, using as ground-truth objects those detected by DETR-DC5-R101~\cite{carion2020end}. We employ class names from MSCOCO~\cite{lin2014microsoft} using the list of synonyms provided in the reference implementation of $\chair$~\cite{rohrbach2018object}. After filtering out all the instances with null $\chair_i$ difference, we end up with 70k and 77k training samples for LLaVA-1.5-7B and LLaVA-MORE-8B, respectively. We refer to Appendix A.1 for more details on the DPO fine-tuning stage, carried out with LoRA~\cite{hu2021lora} for efficiency and performance preservation.

\tit{Datasets and Evaluation Benchmarks}
We evaluate the performance of CHAIR-DPO on three popular hallucination-oriented benchmarks: AMBER~\cite{wang2023amber}, CHAIR-MSCOCO~\cite{yue2024less}, and Object HalBench~\cite{rohrbach2018object, yu2024rlhf}. AMBER is designed to assess the visual hallucination tendencies of MLLMs without relying on external LLMs for response annotation or judgment. Following recent works~\cite{jiang2024modality, wang2024mdpo}, we focus on the generative task, evaluating hallucination proneness using the $\chair_i$, Hallucination Rate (HalRate), and Cognition (Cog) metrics, and assessing object recall via the Coverage (Cover) metric. The $\chair_i$ score is computed as described in Eq.~\ref{eq:chair_i}, while HalRate corresponds to the $\chair_s$ metric~\cite{rohrbach2018object}, defined as the percentage of responses that contain at least one hallucinated object. CHAIR-MSCOCO builds on similar principles but operates on a subset of the MSCOCO dataset~\cite{lin2014microsoft}. It evaluates hallucinations using $\chair_i$ and $\chair_s$, leveraging the predefined MSCOCO object classes as a reference vocabulary and extracting mentioned objects with a standard NLP toolkit, in line with the official $\chair$ implementation~\cite{rohrbach2018object}. Object HalBench also reports $\chair_i$ and $\chair_s$ scores but uses a different subset of the MSCOCO validation set. Unlike CHAIR-MSCOCO, it employs a proprietary LLM (\ie, GPT-3.5) to extract mentioned objects, which improves the precision and recall of object identification. Further details about datasets and metrics are provided in Appendix~\ref{subsec:eval_protocol}.

\begin{table}[t]
  \centering
  \setlength{\tabcolsep}{.2em}
  \resizebox{\linewidth}{!}{
  \begin{tabular}{lccc cccc c cc c cc}
    \toprule
    & & & & \multicolumn{4}{c}{\textbf{AMBER}} & & \multicolumn{2}{c}{\textbf{CHAIR-MSCOCO}} & & \multicolumn{2}{c}{\textbf{Object HalBench}} \\
    \cmidrule{5-8} \cmidrule{10-11} \cmidrule{13-14}
    & \textbf{FT} & \textbf{Ext. Support} & & $\chair_i$ $\downarrow$ &  Cover $\uparrow$ & HalRate $\downarrow$ & Cog $\downarrow$ & &  $\chair_s$ $\downarrow$ & $\chair_i$ $\downarrow$ & & $\chair_s$ $\downarrow$ & $\chair_i$ $\downarrow$ \\
    \midrule
    \rowcolor{lightgray}
    LLaVA-1.5-7B~\cite{liu2023improved} & - & - & & 7.6 & 51.7 & 35.0 & 4.2 & & 50.6 & 13.9 & & . & . \\
    \hspace{0.3cm}+ DoLa~\cite{chuang2024dola} & - & - & & 7.6 & 51.6 & 36.0 & 4.0 & & 51.6 & 14.1 & & - & -  \\
    \hspace{0.3cm}+ VCD~\cite{leng2024mitigating} & - & - & & - & - & - & - & & 48.6 & 14.9 & & 48.0 & 22.3 \\
    \hspace{0.3cm}+ OPERA~\cite{huang2024opera} & - & - & & 7.3 & 49.6 & 32.0 & 3.5 & & 47.8 & 14.6 & & - & -  \\
    \hspace{0.3cm}+ Woodpecker~\cite{yin2024woodpecker} & - & GPT-3.5 & & 6.9 & 48.9 & 30.4 & 3.6 & & 45.8 & 14.8 & & - & - \\
    \hspace{0.3cm}+ POVID~\cite{zhou2024aligning} & \cmark & GPT-4 & & 7.4 & 51.3 & 34.3 & 3.9 & & - & - & &  50.7 & 15.3 \\
    \hspace{0.3cm}+ HA-DPO~\cite{zhao2023beyond} & \cmark & GPT-4 & & 6.7 & 49.8 & 30.9 & 3.3 & & 38.2 & 11.0 & & 39.9 & 19.9  \\
    \hspace{0.3cm}+ HALVA~\cite{sarkar2025mitigating} & \cmark & Gemini & & 6.6 & \underline{53.0} & 32.2 & 3.4 & & 41.4 & 11.7 & & - & - \\
    \hspace{0.3cm}+ EOS~\cite{yue2024less} & \cmark & - & & 5.1 & 49.1 & 22.7 & 2.0 & & 36.8 & 11.3 & & - & -  \\
    \hspace{0.3cm}+ mDPO~\cite{wang2024mdpo} & \cmark & - & & 4.4 & 52.4 & 24.5 & 2.4 & & - & - & & 35.7 & 9.8 \\
    \hspace{0.3cm}+ MFPO~\cite{jiang2024modality} & \cmark & - & & 4.1 & \textbf{55.7} & 22.5 & 1.9 & & - & - & & 13.4 & \underline{6.6} \\
    \hspace{0.3cm}+ REVERSE~\cite{wu2025generate} & \cmark & GPT-4 & & 4.0 & 26.9 & \textbf{10.2} & \textbf{0.9} & & \textbf{13.6} & 6.1 & & - & - \\
    \rowcolor{OurColor}
    \hspace{0.3cm}+ \textbf{CHAIR-DPO$_{(\beta=0.5)}$} & \cmark & - & & 3.8 & 48.7 & 18.4 & 1.8 & & 20.8 & 5.8 & & 20.1 & 10.9 \\
    \rowcolor{OurColor}
    \hspace{0.3cm}+ \textbf{CHAIR-DPO$_{(\beta=0.3)}$} & \cmark & - & & \underline{3.2} & 47.4 & 16.2 & \underline{1.3} & & 16.4 & \underline{4.4} & & \underline{13.1} & 6.7 \\
    \rowcolor{OurColor}
    \hspace{0.3cm}+ \textbf{CHAIR-DPO$_{(\beta=0.2)}$} & \cmark & - & & \textbf{3.0} & 46.6 & \underline{14.7} & \underline{1.3} & & \underline{14.4} & \textbf{3.6} & & \textbf{8.6} & \textbf{4.6} \\
    \midrule
    \rowcolor{lightgray}
    LLaVA-MORE-8B~\cite{cocchi2025llavamore} & - & - & & 8.1 & \textbf{53.2} & 38.4 & 4.0 & & 51.2 & 14.4 & & 50.5 & 24.7 \\
    \hspace{0.3cm}+ DoLa~\cite{chuang2024dola} & - & - & & 7.9 & \underline{53.1} & 38.4 & 4.1 & & 51.8 & 13.8 & & - & -  \\
    \hspace{0.3cm}+ Woodpecker~\cite{yin2024woodpecker} & - & GPT-3.5 & & 7.4 & 50.7 & 36.7 & 3.7 & & 51.0 & 14.3 & & - & - \\
    \hspace{0.3cm}+ REVERSE~\cite{wu2025generate} & \cmark & GPT-4 & & 5.1 & 38.9 & 20.8 & 2.1 & & 25.2 & 8.4 & & - & - \\
    \rowcolor{OurColor}
    \hspace{0.3cm}+ \textbf{CHAIR-DPO$_{(\beta=0.5)}$} & \cmark & - & & 3.4 & 50.7 & 17.1 & \underline{1.2} & & 21.0 & \underline{5.2} & & 19.7 & 10.0 \\
    \rowcolor{OurColor}
    \hspace{0.3cm}+ \textbf{CHAIR-DPO$_{(\beta=0.3)}$} & \cmark & - & & \underline{2.9} & 49.4 & \underline{14.7} & \textbf{1.0} & & \underline{12.3} & 6.2 & & \underline{16.0} & \underline{3.8} \\
    \rowcolor{OurColor}
    \hspace{0.3cm}+ \textbf{CHAIR-DPO$_{(\beta=0.2)}$} & \cmark & - & & \textbf{2.6} & 49.7 & \textbf{14.2} & \textbf{1.0} & & \textbf{11.8} & \textbf{3.1} & & \textbf{9.2} & \textbf{4.5} \\

  \bottomrule
  \end{tabular}
  }
  \vspace{-0.15cm}
  \caption{Comparative evaluation of hallucination mitigation methods for MLLMs, conducted on the AMBER, CHAIR-MSCOCO, and Object HalBench datasets. For each method, we indicate whether the underlying MLLM is fine-tuned (FT) and whether it relies on external support from proprietary LLMs, either via additional training data or post-hoc refinement. Bold-faced and underlined values indicate the best and second-best results.}
  \label{tab:results}
  \vspace{-0.3cm}
\end{table}

\subsection{Experimental Results} 

\tinytit{Comparison with the State of the Art}
We compare in Table~\ref{tab:results} our fine-tuned models against a multitude of recent approaches to alleviate hallucinations in MLLMs. Many of them resort to proprietary MLLMs to collect preference data for fine-tuning~\cite{zhao2023beyond,sarkar2025mitigating,wu2025generate} or design sophisticated data pipelines to craft preferred and dispreferred completions~\cite{jiang2024modality}. Conversely, \ours only requires the offline application of an off-the-shelf detector and the computation of the $\chair_i$, which is widely known in the captioning literature and does not need any engineering. 

We notice that both LLaVA-1.5-7B and LLaVA-MORE-8B combined with \ours always achieve the best results, or the second-best at worst, on metrics directly assessing the hallucination degree. Moreover, we can slightly adapt the behavior of \ours by varying the $\beta$ hyperparameter, which controls the strength of the Kullback-Leibler divergence in DPO (cf. Eq.~\ref{eq:dpo}). Specifically, lower $\beta$ values soften this regularization, achieving lower hallucination rates at the expense of slightly worse Coverage scores on AMBER. This pattern of trading off Coverage for hallucination is consistent in all the considered models. However, \ours manages to strike a good balance between the two metrics, in contrast to other methods. For instance, REVERSE reaches a lower HalRate than $\text{\ours}_{\beta=0.2}$ with LLaVA-1.5-7B on AMBER (10.2 vs 14.7), but its Coverage falls to 26.9, recording a severe -24.8 points drop with respect to the 51.7 points of the baseline model, while for $\text{\ours}_{\beta=0.2}$ the degradation is limited to -5.1 points.

The comparison with LLaVA-MORE-8B is even more favorable to \ours, in that it scores the best hallucination results. Training-free methods that do not modify the weights of LLaVA-MORE-8B, such as DoLa~\cite{chuang2024dola} and Woodpecker~\cite{yin2024woodpecker}, are superior in terms of Coverage, but they greatly fall short against \ours in terms of hallucination metrics.

\begin{table}[t]
  \centering
  \setlength{\tabcolsep}{.25em}
  \resizebox{\linewidth}{!}{
  \begin{tabular}{lcc cccc c cc c cc}
    \toprule
    & & & \multicolumn{4}{c}{\textbf{AMBER}} & & \multicolumn{2}{c}{\textbf{CHAIR-MSCOCO}} & & \multicolumn{2}{c}{\textbf{Object HalBench}} \\
    \cmidrule{4-7} \cmidrule{9-10} \cmidrule{12-13}
    & \textbf{Data Filtering} & & $\chair_i$ $\downarrow$ &  Cover $\uparrow$ & HalRate $\downarrow$ & Cog $\downarrow$ & &  $\chair_s$ $\downarrow$ & $\chair_i$ $\downarrow$ & & $\chair_s$ $\downarrow$ & $\chair_i$ $\downarrow$ \\
    \midrule
    \rowcolor{lightgray}
    LLaVA-1.5-7B~\cite{liu2023improved} & - & & 7.6 & 51.7 & 35.0 & 4.2 & & 50.6 & 13.9 & & 47.4 & 25.6 \\
     \cmidrule(l{0.5cm}){1-13}
    \hspace{0.5cm}+ \textbf{CHAIR-DPO$_{(\beta=0.5)}$} & \xmark & & 4.4 & \textbf{50.1} & 22.8 & 2.0 & & 25.6 & 7.1 & & \textbf{19.7} & 11.1 \\
    \hspace{0.5cm}+ \textbf{CHAIR-DPO$_{(\beta=0.5)}$} & \cmark & & \textbf{3.8} & 48.7 & \textbf{18.4} & \textbf{1.8} & & \textbf{20.8} & \textbf{5.8} & & 20.1 & \textbf{10.9} \\
     \cmidrule(l{0.5cm}){1-13}
    \hspace{0.5cm}+ \textbf{CHAIR-DPO$_{(\beta=0.3)}$} & \xmark & & 4.0 & \textbf{49.0} & 20.2 & 1.4 & & 17.0 & 4.9 & & 16.6 & 8.0 \\
    \hspace{0.5cm}+ \textbf{CHAIR-DPO$_{(\beta=0.3)}$} & \cmark & & \textbf{3.2} & 47.4 & \textbf{16.2} & \textbf{1.3} & & \textbf{16.4} & \textbf{4.4} & & \textbf{13.1} & \textbf{6.7} \\
    \cmidrule(l{0.5cm}){1-13}
    \hspace{0.5cm}+ \textbf{CHAIR-DPO$_{(\beta=0.2)}$} & \xmark & & 3.4 & \textbf{48.4} & 17.2 & \textbf{1.2} & & \textbf{14.0} & \textbf{3.3} & & 11.3 & 5.4 \\
    \hspace{0.5cm}+ \textbf{CHAIR-DPO$_{(\beta=0.2)}$} & \cmark & & \textbf{3.0} & 46.6 & \textbf{14.7} & 1.3 & & 14.4 & 3.6 & & \textbf{8.6} & \textbf{4.6} \\
    \midrule
    \rowcolor{lightgray}
    LLaVA-MORE-8B~\cite{cocchi2025llavamore} & - & & 8.1 & 53.2 & 38.4 & 4.0 & & 51.2 & 14.4 & & 50.5 & 24.7 \\
    \cmidrule(l{0.5cm}){1-13}
    \hspace{0.5cm}+ \textbf{CHAIR-DPO$_{(\beta=0.5)}$} & \xmark & & 3.7 & \textbf{51.5} & 20.1 & 1.7 & & 22.8 & 5.8 & & \textbf{19.6} & \textbf{9.3} \\
    \hspace{0.5cm}+ \textbf{CHAIR-DPO$_{(\beta=0.5)}$} & \cmark & & \textbf{3.4} & 50.7 & \textbf{17.1} & \textbf{1.2} & & \textbf{21.0} & \textbf{5.2} & & 19.7 & 10.0 \\
     \cmidrule(l{0.5cm}){1-13}
    \hspace{0.5cm}+ \textbf{CHAIR-DPO$_{(\beta=0.3)}$} & \xmark & & 3.4 & \textbf{51.0} & 19.0 & 1.4 & & 18.6 & 5.0 & & 14.0 & 7.0 \\
    \hspace{0.5cm}+ \textbf{CHAIR-DPO$_{(\beta=0.3)}$} & \cmark & & \textbf{2.9} & 49.4 & \textbf{14.7} & \textbf{1.0} & & \textbf{16.0} & \textbf{3.8} & & \textbf{12.3} & \textbf{6.2} \\
    \cmidrule(l{0.5cm}){1-13}
    \hspace{0.5cm}+ \textbf{CHAIR-DPO$_{(\beta=0.2)}$} & \xmark & & 2.9 & \textbf{50.1} & 16.4 & 1.4 & & 15.8 & 4.4 & & 12.6 & 6.3 \\
    \hspace{0.5cm}+ \textbf{CHAIR-DPO$_{(\beta=0.2)}$} & \cmark & & \textbf{2.6} & 49.7 & \textbf{14.2} & \textbf{1.0} & & \textbf{11.8} & \textbf{3.1} & & \textbf{9.2} & \textbf{4.5} \\
  \bottomrule
  \end{tabular}
  }
  \vspace{-0.15cm}
  \caption{Ablation study of the application of data filtering. We assess the impact of filtering out any preference instance where the $\chair_i$ difference is zero.}
  \label{tab:ablation}
  \vspace{-0.3cm}
\end{table}

\tit{Ablation Studies}
We assess the impact of filtering out instances where the $\chair_i$ difference between the two candidate completions is null, presenting the results in Table~\ref{tab:ablation}. When data filtering is on, CHAIR-DPO records the best hallucination scores concerning $\chair_i$ and HalRate on AMBER. The same effect is observable on the Cognition metric in AMBER and in general on CHAIR-MSCOCO, except for LLaVA-1.5-7B with $\beta=0.2$, which performs slightly better without the filter. With LLaVA-MORE-8B, the data filtering benefits are consistent on AMBER and CHAIR-MSCOCO, while we notice a minimal degradation on Object HalBench when $\beta$ is set to 0.5. We impute it to the strong regularization imposed by $\beta=0.5$ during the fine-tuning, which limits how much the model can learn from severe changes in the hallucination level measured by $\chair_i$. Finally, we notice the same pattern related to the Coverage metrics on AMBER as in Table~\ref{tab:results}, where models trade off better hallucination rates for lower Coverage.
These results suggest that the proposed data filtering approach is well-suited for \ours, with the added benefit of significantly speeding up fine-tuning by eliminating nearly 90\% of the dataset.

\begin{table}[tb]
  \centering
  \setlength{\tabcolsep}{.35em}
  \resizebox{\linewidth}{!}{
  \begin{tabular}{lc cc c ccc c c c c c c}
    \toprule
    & & \multicolumn{2}{c}{\textbf{MME}} & & \multicolumn{3}{c}{\textbf{SEED}} & & \textbf{MMMU} & & \textbf{Science-QA} & & \textbf{AI2D} \\
    \cmidrule{3-4} \cmidrule{6-8} \cmidrule{10-10} \cmidrule{12-12} \cmidrule{14-14}
    & & Perception & Cognition & & All & Video & Image & & Acc & & Acc & & Acc\\
    \midrule
    InstructBLIP-7B~\cite{dai2023instructblip} & & - & - & & 53.4 & 58.8 & 38.1 & & - & & 60.5 & & - \\
    Qwen-VL-7B~\cite{bai2023qwenvl} & & - & - & & 56.3 & 62.3 & 39.1 & & - & & 67.1  & & - \\
    Qwen-VL-7B-Chat~\cite{bai2023qwenvl} & & 1487.5 & - & & 58.2 & 65.4 & 37.8 & & - & & 68.2 & & - \\
    LLaVA-1.5-LLaMA3-8B~\cite{hanoona2024LLaVA} & & 1544.4 & 330.3 & & 64.3 & 42.0 & 70.1 & & 37.3 & & 74.2 & & 60.7 \\
    \midrule
    LLaVA-1.5-7B~\cite{liu2023improved} & & 1474.3 & 314.6 & & 61.6 & 42.0 & 66.8  & & 34.2 & & 69.0 & & 56.4 \\
    \rowcolor{OurColor}
    \hspace{0.3cm} + \textbf{\ours}$_{(\beta=0.5)}$ && 1520.9 & 372.9 & & 60.8 & 39.5 & 66.4 & & 34.9 & & 69.6 & & 55.0 \\
    \rowcolor{OurColor}
    \hspace{0.3cm} + \textbf{\ours}$_{(\beta=0.3)}$ && 1525.9 & 372.9 & & 60.8 & 39.7 & 66.3 & & 35.0 & & 69.3 & & 54.8 \\
    \rowcolor{OurColor}
    \hspace{0.3cm} + \textbf{\ours}$_{(\beta=0.2)}$ && 1518.8 & 375.0 & & 60.7 & 39.8 & 66.2 & & 35.2 & & 69.3 & & 54.7 \\
    \midrule
    LLaVA-MORE-8B~\cite{cocchi2025llavamore} & &  1531.5 & 353.3 & & 64.1 & 42.4 & 69.8 & & 39.4 & & 76.3 & & 61.8 \\
    \rowcolor{OurColor}
    \hspace{0.3cm} + \textbf{\ours}$_{(\beta=0.5)}$ && 1417.3 & 327.5 & & 64.0 & 43.0 & 69.5 & & 37.1 & & 74.4 & & 59.3 \\
    \rowcolor{OurColor}
    \hspace{0.3cm} + \textbf{\ours}$_{(\beta=0.3)}$ && 1414.0 & 340.7 & & 64.1 & 43.5 & 69.5 & & 36.1 & & 74.4 & & 58.9 \\
    \rowcolor{OurColor}
    \hspace{0.3cm} + \textbf{\ours}$_{(\beta=0.2)}$ && 1412.6 & 335.4 & & 64.1 & 43.8 & 69.5 & & 36.8 & & 74.2 & & 59.0 \\
  \bottomrule
  \end{tabular}
  }
  \vspace{-0.15cm}
  \caption{General cognition evaluation of \ours. We compare LLaVA-1.5-7B and LLaVA-MORE-8B with and without \ours to ensure performance preservation.}
  \label{tab:mllm_benchmarks}
  \vspace{-0.3cm}
\end{table}

\tit{Performance Preservation Analysis}
An important research question is whether \ours hurts the general cognitive capabilities of the reference MLLM, \ie $\piref$ (cf. Eq.~\ref{eq:dpo}). To address this point, we test \ours on several benchmarks typically employed in a comprehensive evaluation of MLLMs~\cite{liu2023improved}. These benchmarks include MME~\cite{fu2023mme}, which measures capabilities in 14 distinct multimodal interaction categories, SEED-Bench~\cite{li2023seed}, which examines 12 cognitive aspects spanning from visual comprehension to text recognition via 19k expert-developed questions, MMMU~\cite{yue2023mmmu}, which challenges models with specialized academic content drawn from collegiate curricula, Science-QA~\cite{lu2022learn}, which probes multidisciplinary reasoning across scientific domains with structured question formats, and AI2D~\cite{kembhavi2016diagram}, which assesses visual-scientific literacy through diagram interpretation exercises.
The results are presented in Table~\ref{tab:mllm_benchmarks}. We highlight that \ours even improves LLaVA-1.5-7B on MME, MMMU, and Science-QA, while scoring comparable to the reference model and other open-source MLLMs on SEED and AI2D. Notably, this behavior is consistent no matter the strength of the $\beta$ regularizer. On the other hand, LLaVA-MORE-8B records a reasonable penalty on most benchmarks, even though \ours enhances its performance on SEED-Video. We argue that this regression on general benchmarks is more than acceptable given the notable reduction in hallucinations testified by Table~\ref{tab:results}. Conversely, we conclude that \ours does not cause a catastrophic forgetting of the knowledge acquired by LLaVA models during visual instruction tuning. We credit for this the Kullback-Leibler regularization of DPO, controlled by $\beta$, as well as the efficient fine-tuning enabled by LoRA, which results in a minimal change to the parameters of the baseline model.

\begin{figure*}[t]
\includegraphics[width=0.985\linewidth]{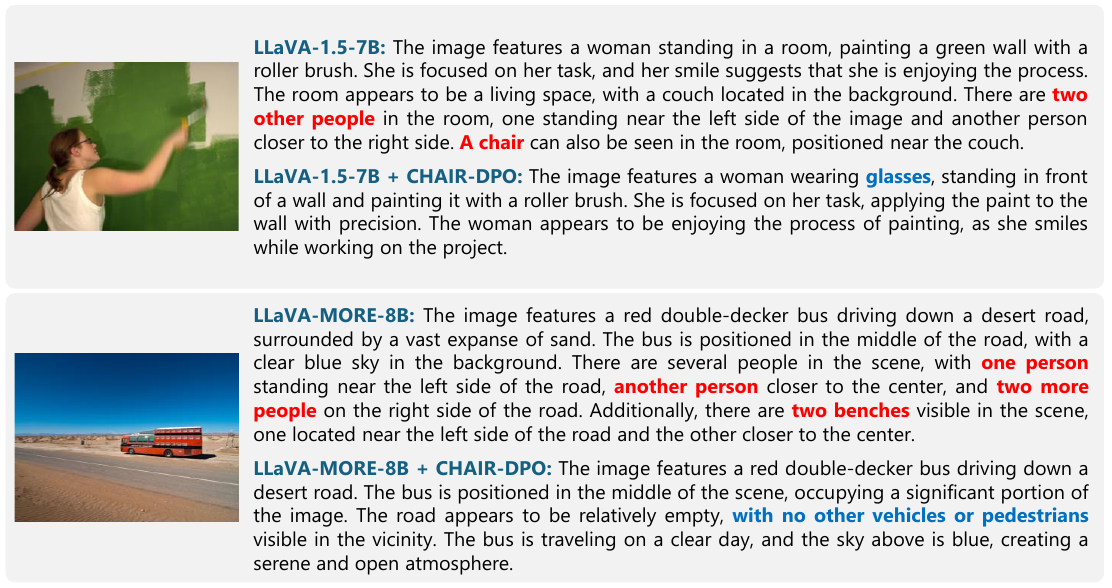}
\vspace{-0.2cm}
\caption{Image captioning qualitative results with LLaVA-1.5-7B and LLaVA-MORE-8B. When applied, \ours not only reduces hallucinations (\ie, \textcolor{OurRed}{\textbf{red words}}), but also incentivizes focusing on items overlooked by the baseline models, outlined in \textcolor{OurBlue}{\textbf{blue}}.}
\label{fig:qualitatives}
\vspace{-0.3cm}
\end{figure*}

\tit{Qualitative Results}
Finally, Figure~\ref{fig:qualitatives} presents captions generated by LLaVA-1.5-7B and LLaVA-MORE-8B~\cite{cocchi2025llavamore} conditioned on the prompt: \texttt{Describe the image}, before and after fine-tuning with \ours. As it can be seen, LLaVA-1.5-7B~\cite{liu2023visual} generates a reasonable description of the first picture, but then mentions the erroneous presence of two other people and a chair in the same room. \ours effectively avoids such hallucinations, and rather adds the detail of the glasses worn by the woman. A similar pattern is repeated with LLaVA-MORE-8B, which hallucinates a total of four alleged people, as well as a pair of benches. Conversely, \ours correctly identifies that no other vehicles nor pedestrian appear in the image other than the double-decker bus, further confirming the effectiveness of the proposed method in mitigating hallucinations. Additional qualitative results of both models are shown in Appendix~\ref{sec:supp_qualitatives}.

\section{Conclusion}
\label{sec:conclusion}
In this work, we introduced CHAIR-DPO, a novel preference optimization method for mitigating hallucinations in Multimodal Large Language Models. By leveraging the well-established $\chair$ metric to build preference data for DPO training, our approach achieves state-of-the-art performance across multiple hallucination benchmarks while requiring only an off-the-shelf object detector in the data collection stage. Unlike existing approaches that rely on complex pipelines or proprietary MLLMs to generate preference data, CHAIR-DPO provides a simpler yet effective alignment framework. Our experiments demonstrate that CHAIR-DPO not only reduces hallucinations significantly but also preserves the general capabilities of the baseline models, with minimal regression on standard benchmarks. The effectiveness of our method suggests that object awareness is a critical component in developing MLLMs that produce factually accurate responses grounded in visual inputs.

\subsection*{Acknowledgments}
We acknowledge the CINECA award under the ISCRA initiative, for the availability of high-performance computing resources. This work has been supported by the EU Horizon projects ``ELIAS'' (GA No. 101120237) and ``ELLIOT'' (GA No. 101214398), by the EuroHPC JU project  ``MINERVA'' (GA No. 101182737), and by the PRIN project ``MUSMA'' (CUP G53D23002930006 - M4C2 I1.1), funded by the EU - NextGenerationEU.

\bibliography{bibliography}

\appendix
\vspace{1cm}

\noindent In the following, we present additional materials about \ours, comprising technical details required to replicate the exact fine-tuning settings, as well as details on the evaluation protocol. Moreover, we show further qualitative results of \ours applied to LLaVA-1.5-7B~\cite{liu2023visual} and LLaVA-MORE-8B~\cite{cocchi2025llavamore}.

\section{Experimental Settings}
\subsection{Additional Implementation Details}
To collect our preference data we sample responses from both models using a temperature of 0.7. For both the full and filtered preference datasets, we hold out 500 samples to serve as a validation set. The final checkpoint is selected based on the lowest $\chair_i$ score observed on this validation set where the score is computed using micro averaging, that is, by aggregating all hallucinated and mentioned objects across the dataset before applying Eq.~1. For training, we employ LoRA with a rank of 128 and an $\alpha$ of 256. We use the Adam~\cite{kingma2014adam} optimizer, along with a cosine learning rate scheduler. The peak learning rate is set to $2\times10^{-6}$, with 33 warmup steps for LLaVA-1.5-7B and 145 warmup steps for LLaVA-MORE-8B. Training is conducted in a distributed, multi-node, multi-GPU environment consisting of 2 nodes with 8 NVIDIA A100 64GB GPUs each. We utilize DeepSpeed ZeRO Stage 2~\cite{rajbhandari2020zero} and gradient checkpointing to optimize memory usage. This setup allows us to use a total batch size of 64 for LLaVA-1.5-7B and 16 for LLaVA-MORE-8B.

\subsection{Evaluation Protocol Details}\label{subsec:eval_protocol}

\tinytit{AMBER}
The AMBER evaluation set comprises 1,004 manually annotated images, each labeled across four dimensions: object existence (visibility), attributes (properties of visible objects), relations (direct contact between objects), and hallucination targets (objects likely to be imagined based on context).
Beyond standard metrics like $\text{CHAIR}_i$ and Hallucination Rate, AMBER includes Coverage, which measures object recall as the ratio of mentioned objects to ground truth objects. It also introduces Cognition, a metric designed to assess whether hallucinations produced by MLLMs align with patterns of human cognition. This is computed as the proportion of hallucinated objects that match the predefined hallucinatory targets.
All AMBER metrics are macro-averaged across the dataset by computing the score independently for each sample and then averaging the per-sample results. For example, $\text{CHAIR}_i$ is computed using Eq.~1 on each image, and the final score is obtained by taking the mean of these values.

\tit{CHAIR-MSCOCO and Object HalBench}
While the original CHAIR benchmark reports results on the Karpathy test split~\cite{Karpathy_2015_CVPR} and a robust test set~\cite{Lu_2018_CVPR}, CHAIR-MSCOCO evaluates model-generated descriptions for 500 images randomly sampled from the MSCOCO validation set. Following the original CHAIR implementation~\cite{rohrbach2018object}, ground-truth object annotations come from COCO ground-truth sentences, while mentioned objects are extracted with an NLP toolkit and preprocessed taking into account plural forms, synonyms, two word compounds to ensure robust evaluation. By contrast, Object HalBench employs a different subsample of MSCOCO validation subset consisting of 300 images. Differently from AMBER, $\chair_i$ and $\chair_s$ scores are aggregated via micro-averaging.

\section{Qualitative Results}
\label{sec:supp_qualitatives}
\begin{figure*}[t]
\includegraphics[width=\linewidth]{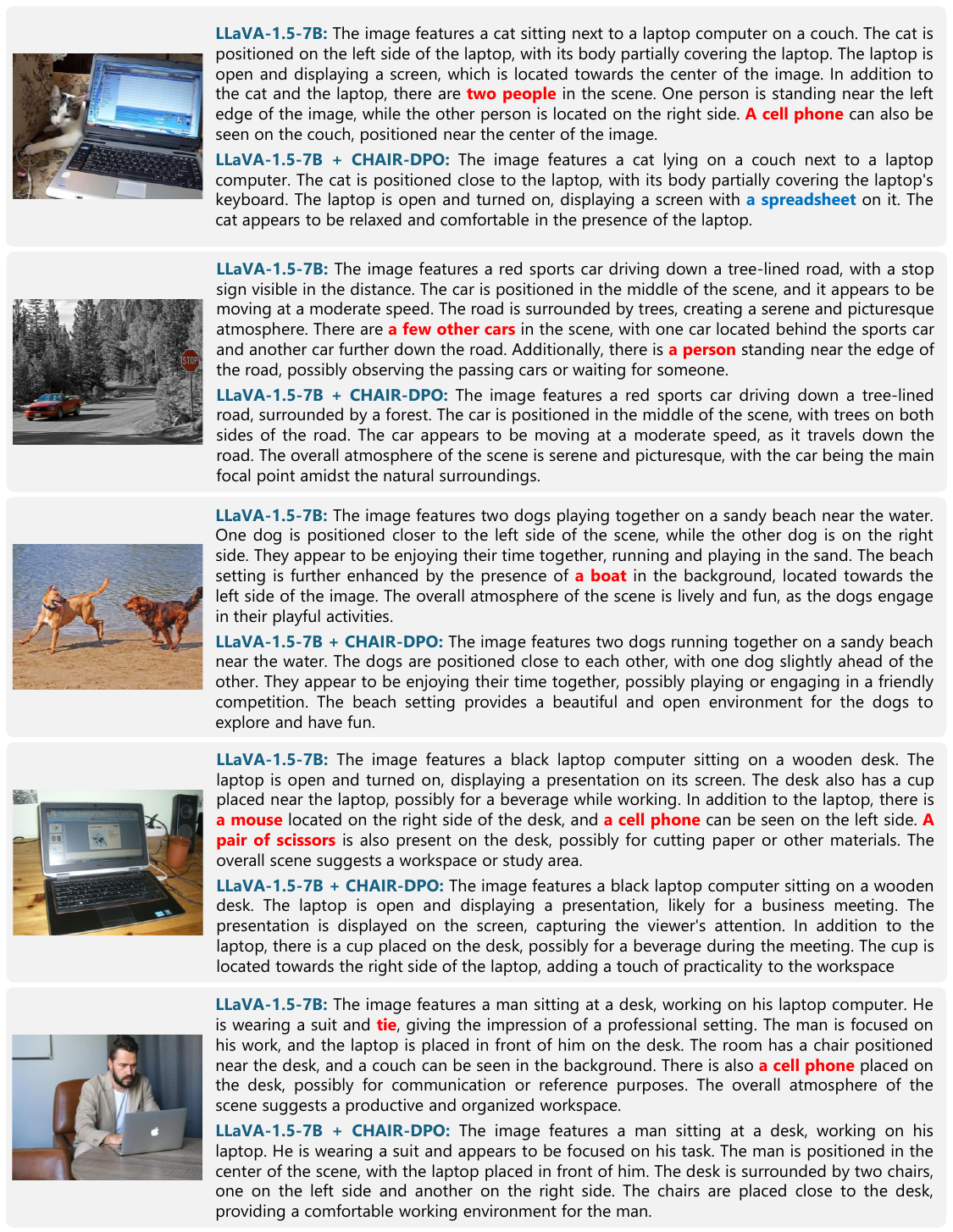}
\vspace{-0.6cm}
\caption{Qualitative results of CHAIR-DPO with LLaVA-1.5-7B~\cite{liu2023improved}. \textcolor{OurRed}{\textbf{Red words}} indicate hallucinations and \textcolor{OurBlue}{\textbf{blue words}} indicate additional details generated by the model.}
\label{fig:qualitatives_suppl_15}
\vspace{-0.3cm}
\end{figure*}

\begin{figure*}[t]
\includegraphics[width=\linewidth]{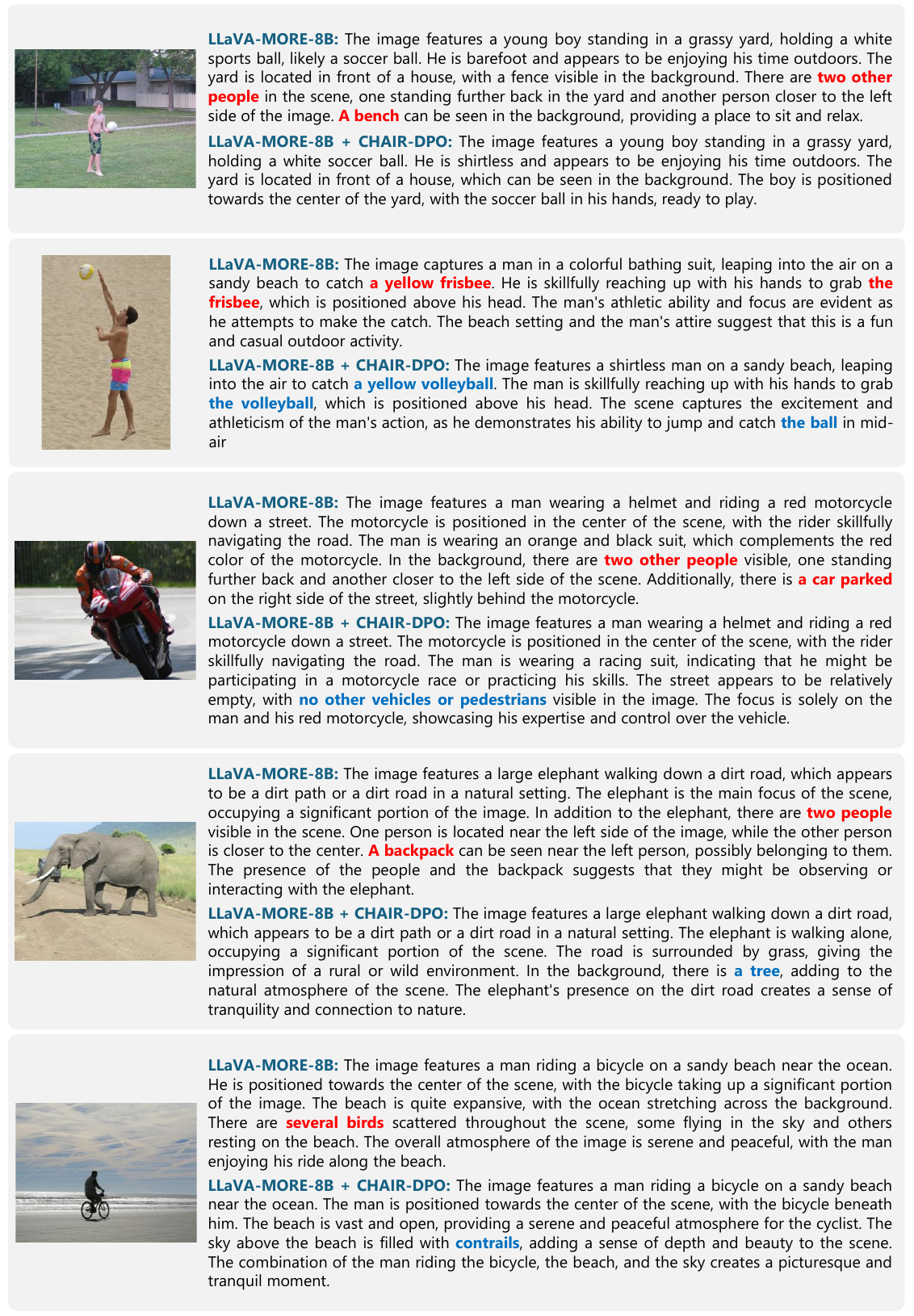}
\vspace{-0.6cm}
\caption{Qualitative results of CHAIR-DPO with LLaVA-MORE-8B~\cite{cocchi2025llavamore}. \textcolor{OurRed}{\textbf{Red words}} indicate hallucinations and \textcolor{OurBlue}{\textbf{blue words}} indicate additional details generated by the model.}
\label{fig:qualitatives_suppl_more}
\vspace{-0.3cm}
\end{figure*}

We show in Figure~\ref{fig:qualitatives_suppl_15} and Figure~\ref{fig:qualitatives_suppl_more} the efficacy of \ours with LLaVA-1.5-7B and LLaVA-MORE-8B respectively. All image descriptions have been generated by conditioning the models with the prompt: \texttt{Describe the image}. Not only \ours reduces visual hallucinations, denoted by the \textcolor{OurRed}{\textbf{red}} font, but also encourages the model to concentrate on additional fine-grained details compared to the baseline, which we outline with in \textcolor{OurBlue}{\textbf{blue}} font. The latter finding especially holds for LLaVA-MORE-8B.

\end{document}